\documentclass[10pt,twocolumn,letterpaper]{article}
\usepackage{fix-cm} 
\usepackage[pagenumbers]{cvpr} 

\definecolor{cvprblue}{rgb}{0.21,0.49,0.74}
\usepackage[pdftex, pagebackref, breaklinks, colorlinks, allcolors=cvprblue]{hyperref}
\usepackage{makecell}
\usepackage{booktabs}

\usepackage{sidecap}
\usepackage{multirow}
\usepackage{xcolor,colortbl}
\usepackage{color, colortbl}
\usepackage{xcolor}
\usepackage{scalerel} 
\usepackage{graphicx} 
\usepackage{array}
\usepackage{booktabs}
\usepackage{colortbl} 
\usepackage{xcolor} 
\usepackage{multirow} 
\usepackage{algorithm}
\usepackage{algpseudocode}
\usepackage{amsmath}
\usepackage{bm}
\usepackage{siunitx}
\newcommand{\bestglobal}{\textsuperscript{$\dagger$}}
\definecolor{tabhighlight}{HTML}{e5e5e5}

\definecolor{tabhighlight}{HTML}{e5e5e5}

\title{Adversarial Prompt Distillation for Vision-Language Models}

\author{
Lin Luo\textsuperscript{1},  
Xin Wang\textsuperscript{1},  
Bojia Zi\textsuperscript{2},  
Shihao Zhao\textsuperscript{3},  
Xingjun Ma\textsuperscript{1}\thanks{Corresponding author: {\tt\small xingjunma@fudan.edu.cn}}, 
Yu-Gang Jiang\textsuperscript{1} \\[0.5em]
\textsuperscript{1}Shanghai Key Lab of Intell. Info. Processing, School of CS, Fudan University \\ 
\textsuperscript{2}The Chinese University of Hong Kong, Shatin, Hong Kong \\ 
\textsuperscript{3}The University of Hong Kong, Pokfulam, Hong Kong \\ 
}

\begin{document}
\maketitle
\begin{abstract}
 Large pre-trained Vision-Language Models (VLMs) such as Contrastive Language-Image Pre-training (CLIP) have been shown to be susceptible to adversarial attacks, raising concerns about their deployment in safety-critical applications like autonomous driving and medical diagnosis. 
One promising approach for robustifying pre-trained VLMs is Adversarial Prompt Tuning (APT), which applies adversarial training during the process of prompt tuning. 
However, existing APT methods are mostly \emph{single-modal} methods that design prompt(s) for only the visual or textual modality, limiting their effectiveness in either robustness or clean accuracy.
In this work, we propose \textbf{Adversarial Prompt Distillation (APD)}, a bimodal knowledge distillation framework that enhances APT by integrating it with multi-modal knowledge transfer. APD optimizes prompts for both visual and textual modalities while distilling knowledge from a clean pre-trained teacher CLIP model. 
Extensive experiments on multiple benchmark datasets demonstrate the superiority of our APD method over the current state-of-the-art APT methods in terms of both adversarial robustness and clean accuracy. The effectiveness of APD also validates the possibility of using a non-robust teacher to improve the generalization and robustness of fine-tuned VLMs.
\end{abstract}    

\section{Introduction}
\label{sec:intro}

Large pre-trained Vision-Language Models (VLMs) ~\cite{radford2021learning, jia2021scaling, li2022blip, li2023blip}  have demonstrated remarkable success in aligning visual and textual representations through a joint embedding space, enabling superior performance on cross-modal tasks. Among these, CLIP~\cite{radford2021learning} stands out by employing contrastive learning to align matched image-text pairs while separating mismatched ones. Despite their effectiveness, these models remain vulnerable to adversarial attacks~\cite{zhang2022towards,lu2023set,he2023sa,han2023ot}, where carefully crafted adversarial perturbations disrupt the image-text alignment, raising concerns for safety-critical applications. 
This pose significant risks on the deployment of pre-trained VLMs in real-world scenarios.

Numerous defense strategies ~\cite{li2024one,zhang2023adversarial,ji2024advlora, zhou2024few, schlarmann2024robust, wang2024pre, mao2022understanding} have been proposed to mitigate adversarial vulnerabilities in VLMs like CLIP, primarily focusing on visual robustness against image attacks. Among these methods, Adversarial Prompt Tuning (APT)  ~\cite{li2024one,zhang2023adversarial,zhou2024few} has been shown to be a promising approach. APT methods apply adversarial training ~\cite{madry2017towards,zhang2019theoretically} during prompt optimization ~\cite{lester2021power,jia2022visual,khattak2023maple} to enhance CLIP's robustness. Two notable APT variants are AdvPT~\cite{zhang2023adversarial}, which defends against fixed pre-generated adversarial images through text prompt tuning but fails against dynamic attacks, and APT~\cite{li2024one}, which uses an adversarially fine-tuned CLIP backbone for improved robustness. However, both methods suffer from limitations: AdvPT's static defense is easily bypassed, while APT relies on a robust backbone. Crucially, their reliance on single-modal (textual) defenses constrains both robustness and clean accuracy.

In this work, we introduce \textbf{Adversarial Prompt Distillation (APD)}, a bimodal defense method designed to enhance the robustness of standard CLIP models on downstream tasks without relying on robustly pre-trained backbones. APD inserts learnable prompts into both visual and textual branches of the model, effectively strengthening cross-modal alignment and  enabling the model to better resist adversarial perturbations. It also employs knowledge distillation~\cite{hinton2015distilling} from a clean teacher CLIP model to improve the adversarial robustness and clean accuracy of the student CLIP model. 
Unlike prior work, APD operates under a more challenging setting where neither the student nor teacher models are robustly-trained, making it broadly applicable to standard pre-trained CLIP models.

We conduct extensive experiments on 11 benchmark datasets to explore the robustness of different variants (textual, visual, or bimodal) of both APT and APD with non-robust CLIP models. We reveal that textual defenses alone offer poor robustness, highlighting their inadequacy against visual attacks.
Visual defenses prove significantly more effective, demonstrating that adversarial robustness can be achieved through visual prompt-based methods. Combining both modalities yields the best robustness, which is further enhanced by distillation from a clean teacher. Overall, APD establishes a strong baseline, achieving superior trade-offs between adversarial robustness and clean accuracy.

In summary, our main contributions are:
\begin{itemize}
    \item We propose \textbf{Adversarial Prompt Distillation (APD)}, a novel defense method that enhances the robustness of standardly trained CLIP models on downstream tasks against adversarial image attacks. APD does not rely on any robustly pre-trained models.
    
    \item APD is a bimodal defense approach that jointly optimizes visual and textual prompts while leveraging knowledge distillation from a clean teacher model to enhance the robustness and accuracy of the student model.
    
    \item Comprehensive evaluations demonstrate that APD outperforms existing APT methods against white-box and transfer attacks, and it also surpasses these methods in terms of domain shift robustness.
\end{itemize}

\section{Related Work}
\label{sec:related works}

\noindent\textbf{Adversarial Attacks on VLMs}\; Adversarial attacks on VLMs can be classified into image-based, text-based, and bimodal attacks. Image-based attacks introduce adversarial perturbations to pixels to corrupt feature representations. In white-box settings, common approaches include FGSM~\cite{goodfellow2014explaining}, PGD~\cite{madry2017towards}, C\&W ~\cite{carlini2017towards} and AutoAttack~\cite{croce2020reliable}, while black-box attacks often rely on surrogate models, with representative methods such as DI-FGSM~\cite{xie2019improving} and TI-FGSM~\cite{dong2019evading}. Text-based attacks manipulate input tokens to disrupt language understanding~\cite{li2020bert, ren2019generating, gao2018black, jin2020bert}. Bimodal attacks, such as Co-Attack~\cite{zhang2022towards}, simultaneously perturb images and text to misalign multimodal embeddings, while SGA~\cite{lu2023set}, SA-Attack~\cite{he2023sa}, and Ot-Attack~\cite{han2023ot} generate transferable adversarial examples across VLMs. This work focuses on defending against white-box image attacks, particularly PGD and AutoAttack.

\vspace{0.1cm}  
\noindent\textbf{Adversarial Prompt Tuning}\;
Prompt tuning has become a popular approach for adapting VLMs~\cite{zhou2022learning, zhou2022conditional, jia2022visual, khattak2023maple, zang2022unified, zhao2024learning, wang2024lion, zhang2022prompting}, with recent extensions to adversarial robustness. Methods like Adversarial Visual Prompting (AVP)~\cite{chen2023visual} employs visual prompting to improve the adversarial robustness of a pre-trained model. TeCoA~\cite{mao2022understanding} and PMG-AFT~\cite{wang2024pre} utilize visual prompt tuning to enhance the zero-shot robustness of CLIP.
On the other hand, AdvPT~\cite{zhang2023adversarial} and APT~\cite{li2024one} apply textual prompt tuning to defend CLIP against image attacks. 
However, the use of static adversarial images in AdvPT limits its effectiveness against dynamic attacks, and APT requires robustly pre-trained backbones. Built upon MaPle~\cite{khattak2023maple}, Few-shot Adversarial Prompt learning (FAP)~\cite{zhou2024few} introduces bimodal tuning to enhance adversarial robustness.
Our work integrates bimodal prompt tuning with knowledge distillation, achieving robustness without relying on pre-trained robust backbones.

\vspace{0.1cm}  
\noindent\textbf{Adversarial Distillation}\; Adversarial distillation improves robustness by transferring knowledge from teacher to student models.
Approaches like Adversarial Robust Distillation (ARD)~\cite{goldblum2020adversarially} align student outputs with teacher clean outputs, while Adversarial Knowledge Distillation (AKD)~\cite{maroto2022benefits} mixes the teacher’s outputs on adversarial examples with clean labels. 
RSLAD~\cite{zi2021revisiting} uses robust soft labels to improve distillation effectiveness. Introspective Adversarial Distillation (IAD) ~\cite{zhu2021reliable} allows the student model to partially trust the teacher through introspection. PeerAiD~\cite{jung2024peeraid} jointly trains student and peer networks on adversarial examples. Unlike prior work, which all relies on a robust teacher, our method combines adversarial prompt tuning with distillation using a standard CLIP teacher, eliminating the need for robust pre-training.

\section{Proposed Method}
\label{sec: methods}

\subsection{Preliminaries}
\noindent\textbf{CLIP Model}\; A typical CLIP model consists of an image encoder $f_v: \mathcal{I} \rightarrow \mathbb{R}^d$, parameterized by $\boldsymbol{\theta}_v$, and a text encoder $f_t : \mathcal{T} \rightarrow \mathbb{R}^d $, parameterized by $\boldsymbol{\theta}_t$. These two encoders respectively extract features from images and texts, mapping inputs from different modalities into unified representations within the joint $d$-dimensional space. For image classification on a dataset $D=\left\{\boldsymbol{x}_i, \boldsymbol{y}_i\right\}_{i=1}^N$ with $C$ classes, CLIP generates textual descriptions $\boldsymbol{t}_j$ using the template $\texttt{"a photo of a <class>"}$ for each class name by default.
For an input image $\boldsymbol{x}_i$ and text $\boldsymbol{t}_j$, their corresponding representations $\boldsymbol{z}_v^{(i)} $ and
$\boldsymbol{z}_t^{(j)}$ can be computed as follows: 
\begin{equation}
\boldsymbol{z}_v^{(i)}=f_v\left(\boldsymbol{x}_i; \boldsymbol{\theta}_v\right), \quad \boldsymbol{z}_t^{(j)}=f_t\left(\boldsymbol{t}_j; \boldsymbol{\theta}_t\right).
\end{equation}
CLIP calculates the similarity between the image representation $\boldsymbol{z}_v^{(i)} $ and the text representation $\boldsymbol{z}_t^{(j)}$ as the logits:
\begin{equation}
q_{i,j} = \cos \left(\boldsymbol{z}_v^{(i)}, \boldsymbol{z}_t^{(j)}\right),
\label{eq:logits}
\end{equation}
 where $\cos (\cdot, \cdot)$ denotes the cosine similarity. The probability that image $\boldsymbol{x}_i$ belongs to the $j$-th class $p_{i,j}$ can be obtained as follows:
\begin{equation}
p_{i,j} =\frac{\exp \left(q_{i,j}  \right)}{\sum_{k=1}^C \exp \left(q_{i,k} \right)}.
\label{eq:probability}
\end{equation}

\vspace{0.1cm} 
\noindent\textbf{Bimodal Prompt Tuning for CLIP}\;
A classic bimodal prompt tuning method is vision-language prompt tuning (VLP)~\cite{khattak2023maple} which tunes both the textual prompts $\boldsymbol{P}_t$ and the visual prompts $\boldsymbol{P}_v$ following the same training schedule. Here, we denote the image and text representation outputs of VLP as:
\begin{equation}
\boldsymbol{z}_v^{(i)}=f_v\left(\boldsymbol{x}_i; \boldsymbol{P}_v , \boldsymbol{\theta}_v\right), \quad \boldsymbol{z}_t^{(j)}=f_t\left(\boldsymbol{t}_j; \boldsymbol{P}_t, \boldsymbol{\theta}_t\right).
\end{equation}
At this point, the logits calculation for the CLIP model, as represented in Eq. \eqref{eq:logits} can be reformulated as follows:
\begin{equation}
q_{i,j} = \cos \left(\boldsymbol{z}_v^{(i)}, \boldsymbol{z}_t^{(j)}\right) = \cos \left( f_v(\boldsymbol{x}_i;\boldsymbol{P}_v,\boldsymbol{\theta}_v),  f_t(\boldsymbol{t}_j; \boldsymbol{P}_t, \boldsymbol{\theta}_t) \right).
\label{eq:logits_vl}
\end{equation}

\subsection{Adversarial Prompt Distillation}
\begin{figure*}[!t]
    \centering
    \includegraphics[width=\textwidth]{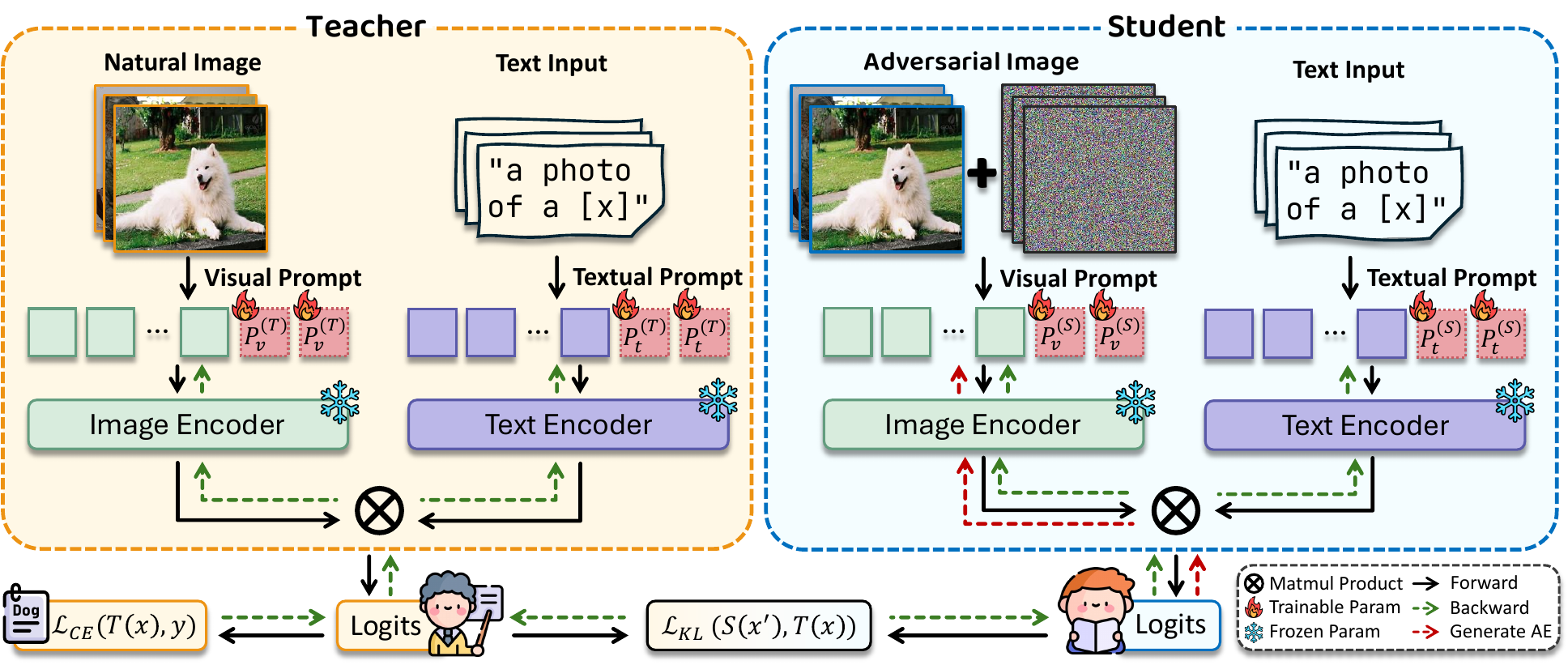}
    \caption{Overview of APD. The student and teacher CLIP are fine-tuned simultaneously using both visual and textual prompts. The teacher model is trained on natural images to achieve high clean performance, while also receiving feedback from the student to help the student better align with the teacher. The student model is trained on adversarial examples and aligns its output with the teacher model.}
    \label{fig:APD overview}
\end{figure*}

\noindent\textbf{Framework Overview}\; 
An overview of our proposed method APD is illustrated in Figure \ref{fig:APD overview}. APD involves two pre-trained CLIP models, designated as the teacher and the student model, respectively. 
The student model takes adversarial images, generated on the fly, as input and learns to align its logits with those of the teacher via the Kullback-Leibler (KL) divergence loss.  
The teacher model processes natural images. During the prompt tuning process, it is tuned to minimize the cross-entropy between its outputs and the ground truth to generate high-quality soft labels. 
At the same time, it receives feedback from the student to assist the student in aligning its logits with that of the teacher. 
Overall, APD defines a bi-level optimization process, involving \emph{inner maximization} to generate adversarial examples and \emph{outer minimization} to tune both the teacher and student via bimodal prompt tuning. Next, we will introduce the inner and outer optimizations in detail.

\vspace{0.1cm} 
\noindent\textbf{Inner Maximization}\;  
During the inner maximization process of APD, adversarial samples are generated on the fly for the student model. Since CLIP processes both text and image data, adversarial examples can be crafted in different modalities: visual, textual, or bimodal. In APD, we focus on visual vulnerability and only generate adversarial examples for the images.
Given an input image $\boldsymbol{x}$, APD generates its adversarial version $\boldsymbol{x}'$ by perturbing the image to maximize the dissimilarity between the image representation and its ground truth text representation (i.e., the representation of the ground truth class prompt).  Formally, this can be achieved by maximizing the cross-entropy loss:
\begin{equation}
\label{eq:generated AE}
\underset{\boldsymbol{x}'}{\operatorname{argmax}} \quad \mathcal{L}_{\textup{CE}}\left(S(\boldsymbol{x'}), \boldsymbol{y}\right),  \;\; \text { s.t. }\left\|\boldsymbol{x}' - \boldsymbol{x} \right\|_{\infty} \leq \epsilon,
\end{equation}
where $S(\boldsymbol{x}')$ denotes the logits output of the student model for adversarial example $\boldsymbol{x}'$.  The logits $S(\boldsymbol{x}')$ can be computed using the similarity function as defined in Eq.~\eqref{eq:logits_vl}. $\mathcal{L}_{CE}$ is the cross-entropy loss, and $\epsilon$ constrains the magnitude of the perturbation.

\vspace{0.1cm} 
\noindent\textbf{Outer Minimization}\; 
During the outer minimization process, APD employs an \textit{online distillation} strategy that simultaneously fine-tunes the teacher and student models.
The teacher model processes only the natural (clean) examples, with its optimization involving two terms: one for natural training and the other for receiving feedback from the student. The corresponding minimization process for the teacher model is formulated as follows:
\begin{equation}
\underset{\boldsymbol{P}^{(T)}}{\operatorname{argmin}} \quad \mathcal{L}_{\textup{CE}}\left(T(\boldsymbol{x}), \boldsymbol{y}\right) + \beta \cdot \mathcal{L}_{KL}\left(T(\boldsymbol{x}), S(\boldsymbol{x}')\right),
\end{equation}
where $T(\boldsymbol{x})$ (can be obtained using Eq.~\eqref{eq:logits_vl}) denotes the logits of the teacher for natural input $\boldsymbol{x}$. The term $\boldsymbol{P}^{(T)} = \{\boldsymbol{P}_v^{(T)}, \boldsymbol{P}_t^{(T)}\}$ represents the updated visual and textual prompts for the teacher. The cross-entropy loss $\mathcal{L}_{CE}$ between the teacher’s logits $T(\boldsymbol{x})$ and the ground truth $\boldsymbol{y}$ aids the teacher in achieving high natural accuracy, ensuring reliable output soft labels. Meanwhile, the KL divergence $\mathcal{L}_{KL}$ quantifies the discrepancy between the outputs of the student model $S(\boldsymbol{x'})$ and the teacher model $T(\boldsymbol{x})$, enabling the teacher to adjust its outputs based on feedback from the student. This feedback assists the student model in more effective training. The hyperparameter $\beta$ balances these two loss terms.

The student model, on the other hand, takes adversarial examples generated by Eq.~\eqref{eq:generated AE} as input, using the soft labels produced by the teacher for supervision. In its optimization, the student model learns robust prompts by minimizing the KL divergence between its probability outputs and the teacher’s soft labels. Formally, the distillation process is formulated as:
\begin{equation}
\underset{\boldsymbol{P}^{(S)}}{\operatorname{argmin}} \quad \mathcal{L}_{KL}\left(S(\boldsymbol{x}'), T(\boldsymbol{x})\right),
\label{eq: student_loss}
\end{equation}
where $\boldsymbol{P}^{(S)} = \{\boldsymbol{P}_v^{(S)}, \boldsymbol{P}_t^{(S)}\}$ represents the updated prompts for the student model. It is important to note that the student's input is an adversarial example $\boldsymbol{x}'$ while the teacher's input is a natural example $\boldsymbol{x}$.
Since the teacher, trained on clean data, provides soft labels with strong generalization, this alignment helps the student become more robust to adversarial perturbations while inheriting the teacher’s generalization capability, thereby achieving a better balance between clean accuracy and
adversarial robustness. The complete procedure of APD is outlined in Algorithm \ref{alg:apd}.

\begin{algorithm}
\caption{Adversarial Prompt Distillation (APD)}
\label{alg:apd}
\small
\begin{algorithmic}[1]
\State \textbf{Requirements:} Pre-trained  CLIP models teacher $T$ and student $S$, Training set $\mathcal{D} $, 
Hyperparameters $\beta$, Learning rate $\eta_t$ and $\eta_s$;

\State \textbf{Output:} Optimized prompts 
$\boldsymbol{P}^{(T)}  $, $\boldsymbol{P}^{(S)} $;

\State Freeze backbones of teacher and student models $\boldsymbol{\theta}^{(T)} $, $\boldsymbol{\theta}^{(S)} $

\State Randomly initialize $\boldsymbol{P}^{(T)} $, $\boldsymbol{P}^{(S)} $

\For{epoch $= 1$ to $N$}
    \For{batch $\{x, y\}$ \textbf{in} $\mathcal{D}$}
        \State {// Step 1. Generate adversarial examples on student}
        \State $\boldsymbol{x}' \gets \text{PGD}(S, \boldsymbol{x}, y)$
        
        \State {// Step 2. Compute logits (using Eq.~\eqref{eq:logits_vl})}
        \State $T(x) \gets \cos \left( f_v(\boldsymbol{x};\boldsymbol{P}^{(T)}_v,\boldsymbol{\theta}^{(T)}_v),  f_t(\boldsymbol{t}; \boldsymbol{P}^{(T)}_t, \boldsymbol{\theta}^{(T)}_t) \right) $
        \State $S(x')  \gets \cos \left( f_v(\boldsymbol{x}';\boldsymbol{P}^{(S)}_v,\boldsymbol{\theta}^{(S)}_v),  f_t(\boldsymbol{t}; \boldsymbol{P}^{(S)}_t, \boldsymbol{\theta}^{(S)}_t) \right) $ 
        
        \State {// Step 3. Compute losses}
        \State $\mathcal{L}_S \gets \mathcal{L}_{\text{KL}}(S(x'), T(x))$
        \State $\mathcal{L}_T \gets \mathcal{L}_{\text{CE}}(T(x), y) + \beta \cdot \mathcal{L}_{\text{KL}}(T(x), S(x'))$
        
        \State {// Step 4. Update teacher and student prompts}
        \State $\boldsymbol{P}^{(S)} \gets \boldsymbol{P}^{(S)} - \eta_s \nabla_{\boldsymbol{P}^{(S)}} \mathcal{L}_S$
        \State $\boldsymbol{P}^{(T)} \gets \boldsymbol{P}^{(T)} - \eta_t \nabla_{\boldsymbol{P}^{(T)}} \mathcal{L}_T$
    \EndFor
\EndFor
\end{algorithmic}
\end{algorithm}

\section{Experiments}
\label{sec: experiment}

\subsection{Experimental Setup}

\noindent\textbf{Datasets and CLIP Models}\; Following previous work~\cite{zhou2024few}, we evaluate our APD on 11 commonly used downstream image classification datasets, including ImageNet~\cite{deng2009imagenet}, Caltech101~\cite{fei2004learning}, Flowers102~\cite{nilsback2008automated}, 
OxfordPets~\cite{parkhi2012cats}, 
FGVCAircraft~\cite{maji2013fine},
Food101~\cite{bossard2014food}, 
StanfordCars~\cite{krause20133d}, 
SUN397~\cite{xiao2010sun}, DTD~\cite{cimpoi2014describing}, EuroSAT~\cite{helber2019EuroSAT} and UCF101~\cite{soomro2012ucf101}. Furthermore, to assess the domain generalization capabilities, we employ four variant datasets of ImageNet: ImageNet-A ~\cite{hendrycks2021natural}, ImageNet-R~\cite{hendrycks2021many}, ImageNet-Sketch~\cite{wang2019learning} and ImageNet-V2~\cite{recht2019imagenet}.
For the CLIP models, we use 2 publicly available variants of CLIP provided by~\cite{radford2021learning}, which adopt ViT-B/16, and ViT-L/14 as their vision components, respectively.

\vspace{0.1cm} 
\noindent\textbf{Defense Baselines}\;
To provide a comprehensive comparison and explore the relationship between visual, textual, and bimodal defenses, we select or construct the following APT-based defense methods as our baselines:

\begin{itemize}

\item \textbf{APT-T}~\cite{li2024one, zhang2023adversarial}  
APT-T is an existing method that uses text prompts to defend against adversarial image attacks. Based on shallow prompting, APT-T learns robust prompts at the input layer in the text modality for defense.

\item \textbf{APT-V}  
We construct APT-V as one of our baselines, which uses visual prompts to defend against adversarial attacks. This method is an adversarial adaptation of Visual Prompt Tuning (VPT~\cite{jia2022visual}), where learnable visual prompts are added to deeper transformer layers of the image encoder.

\item \textbf{APT-VL}  
We construct APT-VL as an additional baseline, which employs bimodal adversarial prompt tuning. This approach combines deep vision and text prompt tuning for robustness, with visual and textual prompts inserted into the transformer layers of the vision and language branches.

\item \textbf{FAP-VL}~\cite{zhou2024few} FAP-VL is an existing method that uses visual and textual prompts for adversarial prompt tuning. It employs a novel training objective to enhance the consistency of bimodal features while encouraging differentiated unimodal features between natural and adversarial examples. 

\end{itemize}

\noindent\textbf{Robustness Evaluation}\;  
In our experiments, we focus exclusively on generating image-based adversarial examples. For robustness evaluation, we evaluate APD's robustness under white-box , black-box, and domain shift scenarios. To ensure a more comprehensive evaluation, we also conduct tests using AutoAttack~\cite{croce2020reliable}, which represents the strongest attack method to date. During training, adversarial examples for both baseline methods and APD were generated using PGD with a perturbation budget of $\epsilon = 1/255$ and a step size of $\alpha = 1/255$ for 2 steps. For evaluation, we use the same perturbation budget and step size as in training, but increase the number of steps to 100 to conduct more thorough testing. In white-box scenarios, we use PGD-100 for testing, while black-box testing employs 100-step TI-FGSM~\cite{dong2019evading} as a transfer attack. In the domain shift evaluation, PGD-100 is applied to test robustness on domain-shifted target datasets. For AutoAttack, we applied a perturbation budget of $\epsilon = 1/255$ to ensure reliable results.

\vspace{0.1cm} 
\noindent\textbf{Implementation Details}\; 
We adopt a few-shot training strategy in all experiments, using 16 shots randomly sampled per class. For APD, we employ two identically sized, cleanly pre-trained CLIP models, with one as the teacher model and the other as the student model. Both models are initialized with the same architecture and pre-trained weights. The prompt length is set to 16 in all experiments, and for deep prompting, the prompt depth is set to 12.  
All models are trained for 50 epochs. For additional training details, we closely follow ~\cite{khattak2023maple}. Specifically, we use an SGD optimizer with a momentum of 0.9. The initial learning rate is set to 0.0035, and we apply a cosine learning rate scheduler with a warm-up strategy in the first epoch. The batch size is 4, and the hyperparameter $\beta$ is 0.2. All experiments were conducted on a single NVIDIA A100 GPU.

\begin{table*}[t]
\centering
\caption{Clean Accuracy (\%) and Robust Accuracy (\%) under PGD-100 attack and TI-FGSM attack for various defense methods across 11 datasets with 16-shots results, using CLIP with ViT-B/16 and ViT-L/14 as the vision backbones. The average results are presented in the last column. The best results are highlighted in \textbf{bold}.
}
\label{tab:main_results}
\resizebox{1.0\linewidth}{!}{
\setlength{\tabcolsep}{0.72mm}{
\begin{tabular}{
  l 
  l 
  l cccccccccccc
}
\toprule
\multicolumn{3}{l}{} & \textbf{ImageNet} & \textbf{Caltech101} & \textbf{DTD} & \textbf{EuroSAT} & \textbf{AirCraft} & \textbf{Flowers} & \textbf{Food101} &  \textbf{OxfordPets} & \textbf{Cars} & \textbf{SUN397} & \textbf{UCF101} & \textbf{Average} \\
\hline
\multirow{15}{*}{\rotatebox{90}{\textbf{ViT-B/16}}} & 
\multirow{3}{*}{{APT-T}} & 
 Clean & \textbf{68.2} & \textbf{94.1} & \textbf{66.5} & 72.2 & 29.4 & 92.4 & \textbf{84.9} & \textbf{90.1} & 69.0 & \textbf{73.6} & \textbf{78.2} & 74.42 \\
 &  & PGD & 1.5 & 23.8 & 6.4 & 0.2 & 0.7 & 2.6 & 0.6 & 2.7 & 0.6 & 1.4 & 2.0 & 3.86 \\
  &  & TI & 23.8 & 66.5 & 27.5 & 7.3 & 12.1 & 52.6 & 26.0 & 48.0 & 28.2 & 27.2 & 28.4 & 31.60 \\

    \cmidrule{2-15}
 & \multirow{3}{*}{{APT-V}} & 
Clean & 62.8 & 94.0 & 60.5 & 72.4 & 46.0 & 94.0 & 73.7 & 88.3 & 74.5 & 68.2 & 75.4 & 73.62 \\
 &  & PGD & 27.0 & 75.2 & 36.5 & 27.6 & 21.9 & 72.5 & 23.0 & 50.9 & 34.3 & 31.8 & 36.2 & 39.72\\
 &  & TI & 47.9 & 88.6 & 52.2 & 52.3 & 37.7 & 87.2 &  48.4 & 72.6 & 62.1 & 54.9 & 58.6 & {60.29} \\

     \cmidrule{2-15}
 & \multirow{3}{*}{{APT-VL}} & 
Clean & 61.0 & 93.8 & 62.7 & 36.0 & \textbf{51.0} & 95.5 & 66.8 & 87.4 & 81.7 & 68.3 & 76.6 & 70.98\\
 &  & PGD  & 27.5 & 78.7 & 38.1 & 30.8 
 & 24.2 & 76.8 & 27.8 & 52.6 
 & 46.6 & 34.5 & 44.0 & 43.78 \\
 &  & TI & 46.5 & 88.1 & {52.6} & 32.3 & 39.0 & 88.9 & 46.9 & {74.2} & {70.3} & 55.5 & 60.8 & 59.55 \\

   \cmidrule{2-15}
  & \multirow{3}{*}{FAP-VL} & 
Clean & 61.6 & 93.8 & 62.2 & \textbf{85.8} &  50.3 & \textbf{95.6} & 67.5 & 86.5 & \textbf{81.9} & 68.7 & 74.8 & 75.34\\
 &  & PGD  & {28.7} & {79.1} & {40.1} & {33.6} 
 & 24.5 & 78.0 & 28.2 & 54.6 
 & \textbf{49.9} & 36.2 & 44.8 & 45.25\\
 &  & TI & 47.5 & {89.0} & 40.1 & 44.1 & {39.3} & {89.0} & 48.1 & 73.7 & \textbf{70.7} & {55.8} & {61.4} & 59.88 \\

  \cmidrule{2-15}
 & \multirow{3}{*}{{APD(Ours)}} & 
\cellcolor{gray!20}Clean & \cellcolor{gray!20}{64.9} & \cellcolor{gray!20}93.9 & \cellcolor{gray!20}{65.5} & \cellcolor{gray!20}{76.9} & \cellcolor{gray!20}47.8 & \cellcolor{gray!20}94.4 & \cellcolor{gray!20}71.5 & \cellcolor{gray!20}88.5 & \cellcolor{gray!20}81.1 & \cellcolor{gray!20}71.4 & \cellcolor{gray!20}77.1 & \cellcolor{gray!20}\textbf{75.73}\\
 &  & \cellcolor{gray!20}PGD  &  \cellcolor{gray!20}\textbf{29.4} & \cellcolor{gray!20}\textbf{80.1} &
 \cellcolor{gray!20}\textbf{44.7} & \cellcolor{gray!20}\textbf{47.3} & \cellcolor{gray!20}\textbf{25.4} & \cellcolor{gray!20}\textbf{79.0} & \cellcolor{gray!20}\textbf{33.7 }& \cellcolor{gray!20}\textbf{56.5} & \cellcolor{gray!20}{48.2} & \cellcolor{gray!20}\textbf{37.8 }& \cellcolor{gray!20}\textbf{50.1} & \cellcolor{gray!20}\textbf{48.38}\\
 &  & \cellcolor{gray!20}TI & \cellcolor{gray!20}\textbf{49.9} & \cellcolor{gray!20}\textbf{89.1} & \cellcolor{gray!20}\textbf{56.6} & \cellcolor{gray!20}\textbf{58.1} & \cellcolor{gray!20}\textbf{39.4} & \cellcolor{gray!20}\textbf{89.5} & \cellcolor{gray!20}\textbf{53.3 }& 
 \cellcolor{gray!20}\textbf{76.0} & \cellcolor{gray!20}68.5 & \cellcolor{gray!20}\textbf{57.8 }& \cellcolor{gray!20}\textbf{63.2} & \cellcolor{gray!20}\textbf{63.76} \\

\midrule \noalign{\vspace{-1.0pt}} \midrule

\multirow{15}{*}{\rotatebox{90}{\textbf{ViT-L/14}}} & 
\multirow{3}{*}{{APT-T}} & 
Clean & \textbf{77.3} & \textbf{97.1} & \textbf{75.2} & 83.0 & 47.4 & \textbf{97.5}  & \textbf{91.3} & \textbf{93.9} & 83.8 & \textbf{79.0} & \textbf{86.3} & \textbf{82.89}\\
 &  & PGD & 16.9 & 75.8 & 44.0 & 29.6 & 6.5 & 63.5 & 29.3 & 38.1 & 38.0 & 37.1 & 39.5 & 38.03\\
 &  & TI & 51.3 & 90.8 & 60.6 & 38.9 & 33.5 & 90.0 & 63.3 & 84.2 & 67.7 & 63.7 & 66.0 & 64.55 \\

 \cmidrule{2-15}
 & \multirow{3}{*}{{APT-V}} & 
Clean & 71.6 & 96.7 & 63.9 & 59.6 & 42.9 & 92.5 & 83.3 & 92.9 & 77.6 & 72.2 & 78.1 & 75.57\\
 &  & PGD & 44.1 & 85.6 & 36.8 & 28.8 & 18.3 & 69.6 & 42.1 & 68.9 & 43.5 & 45.5 & 50.2 & 48.49\\
 &  & TI & 60.0 & 92.8 & 52.4 & 46.6 & 32.6 & 83.6 & 64.2 & 84.9 & 65.2 & 61.2 & 64.1 & 64.33 \\

   \cmidrule{2-15}
 & \multirow{3}{*}{{APT-VL}} & 
Clean & 69.9 & 96.5 & 66.1 & 66.3 & \textbf{56.3 }& 96.3 & 77.7 & 92.0 & 86.9 & 73.5 & 81.9 & 78.49\\
 &  & PGD  & 54.4 & 93.1 & 61.3 & 59.3 
 & 38.0 & 93.5 & 66.8 & 84.9 
 & 69.3 & 66.4 & 74.1 & 69.19\\
  &  & TI & 57.9 & 94.8 & 61.9 & 61.8 & 49.9 & 94.1 & 67.5 & 87.0 & 80.0 & 68.9 & 76.0 & 72.71 
 \\

  \cmidrule{2-15}
  & \multirow{3}{*}{{FAP-VL}} & 
Clean & 71.1 & 96.1 & 68.1 & 62.1 & \textbf{56.3} & 96.7 & 77.9 & 92.2 & \textbf{87.0} & 74.5 & 83.3 & 78.66\\
 &  & PGD  & 55.2 & 93.1 & 63.4 & 56.6 
 & 38.9 & 93.8 & 67.4 & 87.4 
 & 70.0 & \textbf{67.4} & 76.3 & 69.95\\
 &  & TI & 57.3 & 94.3 & 63.4 & 58.7 & 50.3 & 94.2 & 67.7 & 87.9 & 80.3 & 68.9 & 77.3 & 72.75 \\

 \cmidrule{2-15}
 & \multirow{3}{*}{{APD(Ours)}} & 
 \cellcolor{gray!20}Clean & \cellcolor{gray!20}73.2 & \cellcolor{gray!20}97.0 & \cellcolor{gray!20}72.3 & \cellcolor{gray!20}\textbf{83.3} & \cellcolor{gray!20}55.1 & \cellcolor{gray!20}97.2 & \cellcolor{gray!20}81.6 & \cellcolor{gray!20}92.6 & \cellcolor{gray!20}84.7 & \cellcolor{gray!20}76.7 & \cellcolor{gray!20}83.7 & \cellcolor{gray!20}81.58\\
 &  & \cellcolor{gray!20}PGD  & \cellcolor{gray!20}\textbf{58.1} & \cellcolor{gray!20}\textbf{94.4} & \cellcolor{gray!20}\textbf{68.9} & \cellcolor{gray!20}\textbf{73.4}
 & \cellcolor{gray!20}\textbf{46.6} & \cellcolor{gray!20}\textbf{95.0} & \cellcolor{gray!20}\textbf{68.0} & \cellcolor{gray!20}\textbf{87.8} 
 & \cellcolor{gray!20}\textbf{72.1} & \cellcolor{gray!20}\textbf{67.4} & \cellcolor{gray!20}\textbf{77.1} & \cellcolor{gray!20}\textbf{73.53}\\
  &  & \cellcolor{gray!20}TI & \cellcolor{gray!20}\textbf{62.9} & \cellcolor{gray!20}\textbf{95.5} & \cellcolor{gray!20}\textbf{70.5} & \cellcolor{gray!20}\textbf{77.3} & \cellcolor{gray!20}\textbf{50.8} & \cellcolor{gray!20}\textbf{95.6} & \cellcolor{gray!20}\textbf{70.4} & 
 \cellcolor{gray!20}\textbf{89.3} & \cellcolor{gray!20}\textbf{80.6} & \cellcolor{gray!20}\textbf{72.5} & \cellcolor{gray!20}\textbf{79.3} & \cellcolor{gray!20}\textbf{76.79}\\

\bottomrule
\end{tabular}}}
\end{table*}

\subsection{Main Results} We evaluate the robustness of APD method under both white-box and black-box attack settings. In addition, we conduct experiments using AutoAttack~\cite{croce2020reliable} , a state-of-the-art attack method, to assess the robustness of APD against the most challenging scenarios. Furthermore, we investigate the defense's performance under domain shift conditions, exploring its robustness across different data distributions. We present and discuss these results as follows.

\vspace{0.1cm} 
\noindent\textbf{White-box Robustness} In the white-box setting, we evaluate the adversarial robustness of APD using the widely adopted PGD~\cite{madry2017towards} attack.
Table \ref{tab:main_results} presents the results of the 4 baseline methods (APT-T, APT-V, APT-VL and FAP-VL) and our APD across 11 datasets. We report the average performance in terms of clean accuracy and adversarial robustness against PGD-100 attack in the last column. 
As can be observed, our APD method demonstrates consistent robustness improvements across different CLIP vision backbones. Specifically, our APD demonstrates superior robustness to all baseline methods on 10 out of 11 datasets when evaluated on CLIP with a ViT-B/16 backbone, and on all 11 datasets with a ViT-L/14 backbone.
Notably, with the ViT-B/16 backbone, our approach surpasses the strongest baseline, FAP-VL, by an average margin of 3.13\% in adversarial robustness. This advantage becomes more pronounced with the larger ViT-L/14 backbone, where APD achieve a 3.58\% improvement over FAP-VL. In terms of clean accuracy, when using the ViT-B/16 backbone, our APD method achieves the highest performance among all baseline methods on average. On the ViT-L/14 backbone, APD ranks second in clean accuracy (merely 1.31\% lower than APT-T), which remains highly competitive, especially considering that the baseline method APT-T with the highest clean accuracy has a significantly lower robustness than APD (38.03\% v.s. 73.53\%). These results collectively demonstrate that APD successfully balances accuracy and robustness, underscoring its practical value in robust vision-language modeling.

From a modality perspective, relying solely on unimodal textual prompt defenses -- as in APT-T -- leads to poor robustness, particularly when the vision backbone itself is less robust. For instance, when using ViT-B/16 as the vision backbone, APT-T achieves only 3.86\% average robustness. Even with the more robust ViT-L/14 backbone, APT-T still shows a significant performance gap compared to bimodal defenses, falling short of our APD by a large margin of 35.5\% in robustness. In contrast, the unimodal visual defense method APT-V consistently outperforms its textual counterpart (APT-T) across both backbones. However, it still lags behind bimodal approaches. Specifically, with ViT-B/16 as vision backbone, APT-V's robustness is on average 8.66\% lower than that of APD, and this gap widens to 25.04\% on ViT-L/14. These results highlight the necessity of incorporating both visual and textual modalities for effective defense,  especially in larger models. Among bimodal defense methods, our APD generally outperforms the other two baselines (APT-VL and FAP-VL) in terms of both robustness and clean accuracy across two different vision backbones, demonstrating strong overall superiority.

\vspace{0.1cm} 
\noindent\textbf{{Black-box Robustness}}
We evaluate the black-box robustness of our APD method using the TI-FGSM~\cite{dong2019evading} attack. The evaluation results are also summarized in Table~\ref{tab:main_results}. It is evident that all defense methods generally exhibit stronger robustness against black-box attacks compared to white-box attacks. Under the black-box setting, our APD method achieves state-of-the-art performance with two different visual backbones (ViT-B/16 and ViT-L/14) across nearly all datasets, with the exception of the StanfordCars~\cite{krause20133d} dataset when using ViT-B/16 as the backbone. On average, APD outperforms the best baseline methods by 3.47\% and 4.04\% on ViT-B/16 and ViT-L/14, respectively, further demonstrating the superior robustness of our APD in black-box scenarios.

\begin{figure}
    \centering
    \includegraphics[width=\columnwidth]{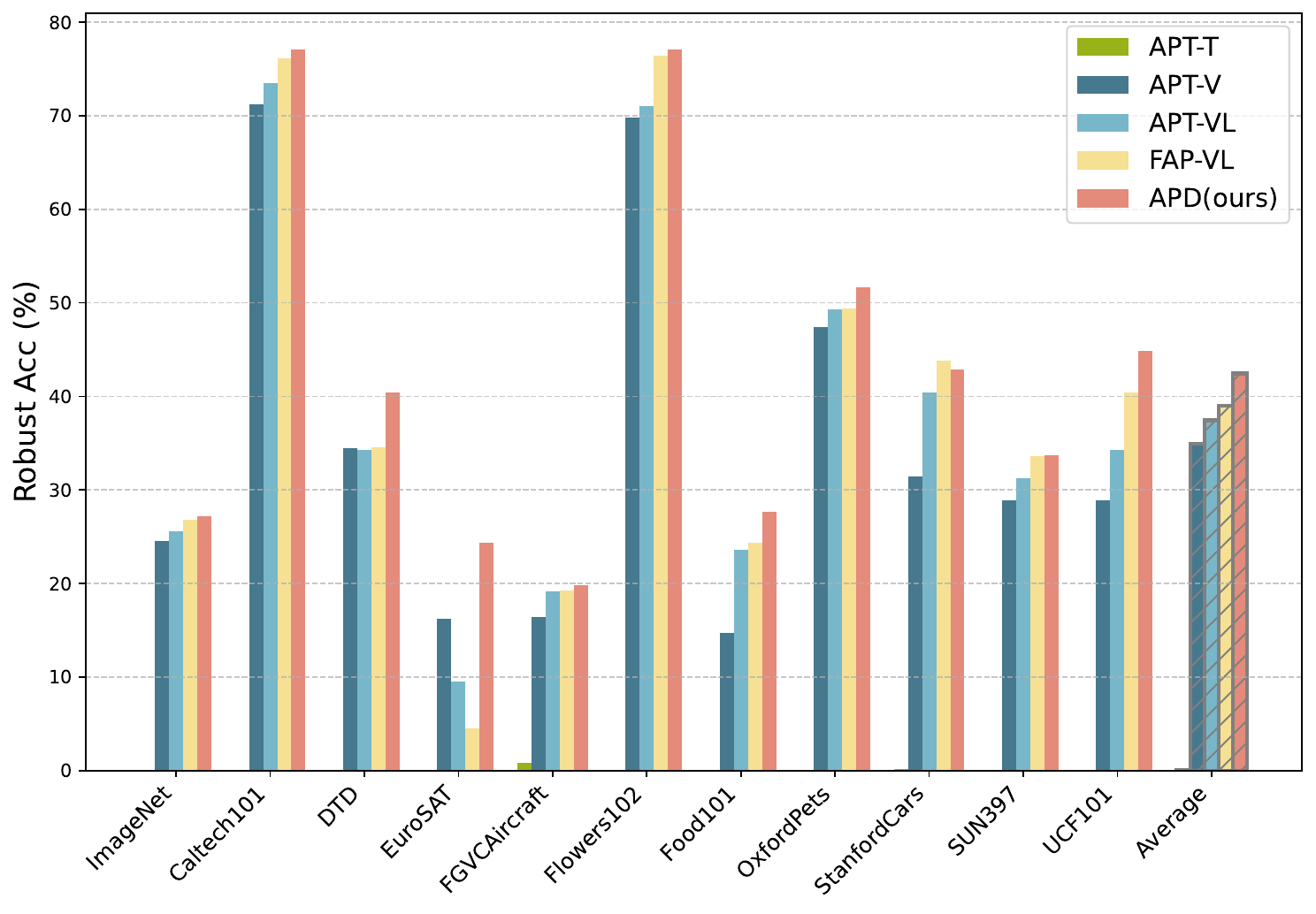}
    \caption{Adversarial robustness (\%) of different defense methods against Auto-Attack on 11 datasets.The last group represents the average performance across all datasets.}
    \label{fig:AutoAttack}

\end{figure}

\vspace{0.1cm} 
\noindent\textbf{AutoAttack Robustness}
We also evaluate our APD method against the AutoAttack, which is generally recognized as the strongest image attack to date. The evaluation was conducted using ViT-B/16 as the vision backbone. The results are presented in Figure \ref{fig:AutoAttack}.  Compared to the PGD-100 results reported in Table~\ref{tab:main_results}, all methods experience a significant drop in robustness, which is expected given the increased strength of AutoAttack. Nevertheless, consistent with the results in Table~\ref{tab:main_results}, our APD method achieves the best AutoAttack robustness on 10 out of 11 datasets.
 While there is a slight gap between our APD and FAP-VL on the StanfordCars~\cite{krause20133d} dataset, our method demonstrates clear advantages on datasets such as DTD~\cite{cimpoi2014describing}, EuroSAT~\cite{helber2019EuroSAT} and UCF101~\cite{soomro2012ucf101}. On average, APD achieves 3.42\% higher robustness than the strongest baseline method, FAP-VL. These results highlight the effectiveness of APD even under the most challenging attack conditions, further demonstrating its robust performance against strong adversarial threats.
It is worth noting that the baseline method APT-T, which relies solely on single-modality textual prompts for defense, exhibits almost no defense capability against AutoAttack, with near-zero AutoAttack robustness on many datasets. This demonstrates that relying solely on textual defenses is highly inadequate and easily circumvented.

\vspace{0.1cm} 
\noindent\textbf{{Domain Shift Robustness}}
To evaluate the domain shift robustness of APD and four baseline methods, we conduct tests on four domain-shifted targeted datasets: ImageNet-A ~\cite{hendrycks2021natural}, ImageNet-R~\cite{hendrycks2021many}, ImageNet-Sketch~\cite{wang2019learning} and ImageNet-V2~\cite{recht2019imagenet}, using ImageNet~\cite{deng2009imagenet} as the source domain. As shown in Table \ref{tab:domain_shift},  our APD method consistently outperforms all baseline methods on both ViT-B/16 and ViT-L/14 backbones across all four datasets. Specifically, with the ViT-B/16 as the vision backbone, APD achieves an average robustness of 18.68\% over the four domain-shifted datasets, surpassing all baselines. This advantage amplifies substantially with the larger ViT-L/14 backbone, where APD attains an average robustness of 43.18\%, outperforming the best baseline FAP-VL by 2.6\%. The superior robustness on ViT-L/14 can be attributed to the enhanced model capacity, which allows APD to learn more informative prompts, thereby more effectively correcting adversarial perturbations. Critically, these advancements require no robust pre-training, ensuring seamless integration into existing vision-language models.

\begin{table}[t]
\centering
\caption{Comparison of domain shift robustness across different defense methods under PGD-100 attack. The best results are highlighted in \textbf{bold}. 
}
\label{tab:domain_shift}
\resizebox{\linewidth}{!}{ 
\setlength{\tabcolsep}{3.5pt}
\renewcommand{\arraystretch}{1.35} 
\fontsize{25}{22}\selectfont 
\begin{tabular}{llcccc}
\toprule
\multirow{2}{*}{} & \multirow{2}{*}{} & \multicolumn{4}{c}{\textbf{\fontsize{25}{25}\selectfont Domain Shift Robustness (\%)}} \\
\cmidrule{3-6}
& & \scalebox{0.9} {ImageNet-A} & \scalebox{0.9} {ImageNet-R} & \scalebox{0.9} {ImageNet-Sketch} & \scalebox{0.9} {ImageNet-V2} \\
\midrule
\multirow{5}{*}{\raisebox{-0.7em}{\rotatebox[origin=c]{90}{\fontsize{25}{25}\selectfont \textbf{ViT-B/16}}}} 
& {APT-T}    & 0.2 & 6.4 & 6.9 & 0.9 \\
& {APT-V}    & \textbf{1.4} & 31.2 & 18.9 & 19.9 \\
& {APT-VL}   & 1.1 & 26.2 & 15.1 & 20.2 \\
& {FAP-VL}      & 1.2 & 27.1 & 16.1 & 21.0 \\
& \cellcolor{gray!20}{APD(Ours)} & \cellcolor{gray!20}\textbf{1.4} & \cellcolor{gray!20}\textbf{31.5} & \cellcolor{gray!20}\textbf{19.1} & \cellcolor{gray!20}\textbf{22.7} \\
\cmidrule{1-6}
\multirow{5}{*}{\raisebox{-0.7em}{\rotatebox[origin=c]{90}{\fontsize{25}{25}\selectfont \textbf{ViT-L/14}}}}
& {APT-T}    & 2.8 & 27.1 & 20.8 & 13.0\\
& {APT-V}    & 9.2 & 58.0 & 39.7 & 36.0 \\
& {APT-VL}   & 8.5 & 52.6 & 34.9 & 43.6 \\
& {FAP-VL}      & 14.5 & 59.2 & 39.6 & 49.0 \\
& \cellcolor{gray!20}{APD(Ours)} & \cellcolor{gray!20}\textbf{15.1} & \cellcolor{gray!20}\textbf{63.0} & \cellcolor{gray!20}\textbf{43.1} & \cellcolor{gray!20}\textbf{51.5} \\
\bottomrule
\end{tabular}
}
\vspace{-0.45cm}
\end{table}

\section{Ablation Studies}

\begin{figure*}[!t]
    \centering
    \includegraphics[width=\textwidth]{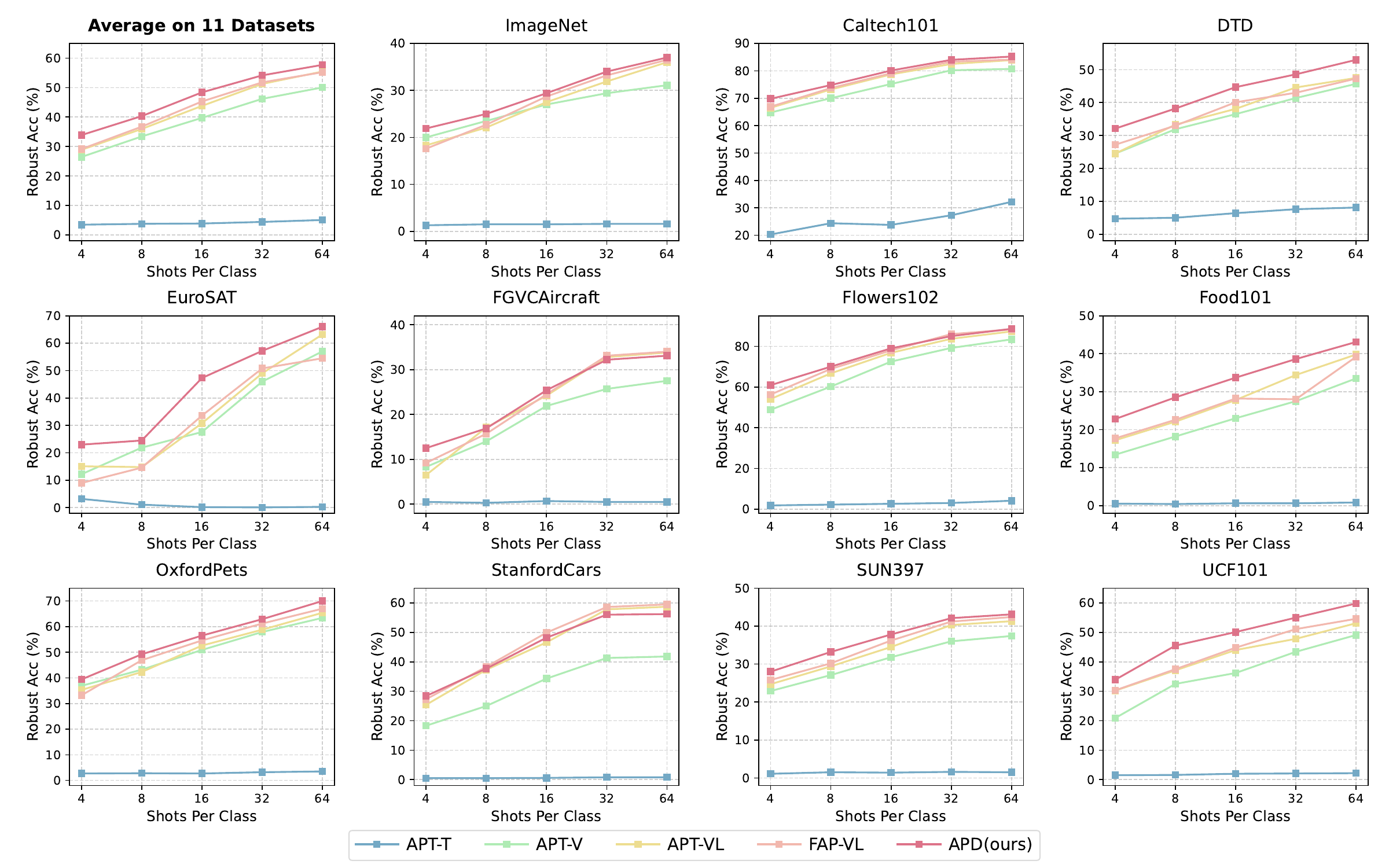}
    \caption{Robustness results (PGD-100) of APD and the other four baseline methods under different shot settings, evaluated with CLIP using ViT-B/16 as the vision backbone.}
    \label{fig:different_shots}
\end{figure*}

\noindent\textbf{Different Shots} We first examine the effect of the number of shots on the robust accuracy of APD and four baseline methods. The evaluation was conducted under different few-shot settings \{4, 8, 16, 32, 64\}, using ViT-B/16 as the vision backbone. The results are presented in Figure \ref{fig:different_shots}. Overall, as the number of shots increases, the robustness of all methods improves, with the exception of APT-T, which exhibits consistently poor robustness. Despite the increase in shots, APT-T shows minimal improvement in robustness, indicating its limited ability to benefit from additional training data. This highlights its inherent weakness in defense performance when applied to a non-robust backbone. On average, our method generally outperforms all baseline methods across all shot settings, demonstrating its effectiveness across varying few-shot settings.

\begin{figure}
    \centering
    \includegraphics[width=\columnwidth]{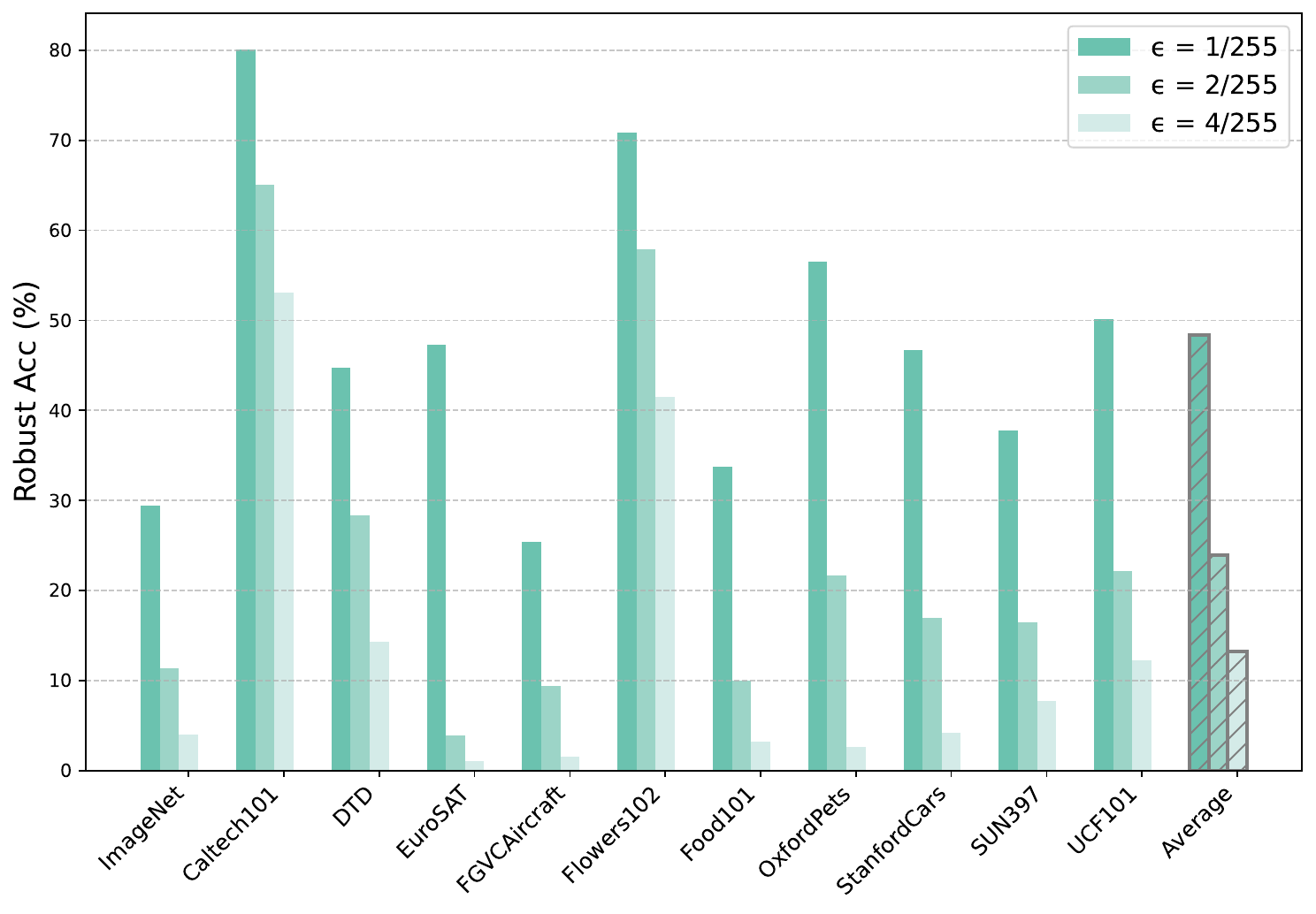}
    \caption{The PGD-100 robustness performance of our APD under different perturbation bounds. The last group represents the average performance across all datasets.
    \vspace{-0.45cm}
}
    \label{fig:eps}

\end{figure}

\vspace{0.1cm} 
\noindent\textbf{Different Perturbation Bounds} To assess the performance of APD under varying attack strengths, we evaluate its robustness under different perturbation bounds. Specifically, we conduct experiments with $\epsilon \in \{1/255, 2/255, 4/255\}$, applying the same $\epsilon$ value during both training and testing. As illustrated in Figure \ref{fig:eps}, increasing the perturbation budget strengthens the adversarial attack, leading to a sharp decline in robust performance.  
This performance drop is also observed in APT ~\cite{li2024one,zhou2024few}. Since prompt tuning is a lightweight adaptation approach, achieving strong robustness under large perturbations typically requires more training data ~\cite{wang2024revisiting, zhai2019adversarially}.

\vspace{0.1cm}  
\noindent\textbf{Unimodal Defense vs. Bimodal Defense} We compare the robustness of unimodal and bimodal defense methods across 11 datasets, using CLIP with ViT-B/16 as the vision backbone. Specifically, we focus on three groups: defense using only textual prompts (Unimodal-T), defense using only visual prompts (Unimodal-V), and defense using both textual and visual prompts (Bimodal-VL). Each of these groups includes two methods for comparison: Adversarial Prompt Tuning (APT) and Adversarial Prompt Distillation (APD). As shown in Table \ref{tab: uni_bi}, the bimodal defenses consistently outperform their unimodal counterparts under both APT and APD settings. Specifically, in the APT framework, the bimodal defense APT-VL achieves 39.92\% and 4.06\% higher robustness compared to its unimodal counterparts APT-T and APT-V, respectively. Similarly, in the APD framework, our bimodal approach APD-VL(Ours) improves over the unimodal methods APD-T and APD-V by 44.27\% and 6.05\%.  These consistent gains highlight the critical role of integrating both modalities in achieving robust defenses. Furthermore, within each group, a clear advantage of APD over APT can be observed, demonstrating the effectiveness of online prompt distillation in enhancing robustness. 

\begin{table}[ht]
\caption{Robustness comparison between unimodal and bimodal defense methods across 11 datasets. Three groups (Unimodal-T, Unimodal-V, Bimodal-VL) are compared with APT and APD. Best methods within each group are marked with \textsuperscript{*}, and the overall best method is marked with \bestglobal.}
\centering
\resizebox{\linewidth}{!}{  
\begin{tabular}{@{}>{\footnotesize}lSSSSSS@{}} 
\toprule
\multirow{2}{*}{} & 
\multicolumn{2}{c}{\textbf{Unimodal-T}} & 
\multicolumn{2}{c}{\textbf{Unimodal-V}} & 
\multicolumn{2}{c}{\textbf{Bimodal-VL}} \\
\cmidrule(lr){2-3} \cmidrule(lr){4-5} \cmidrule(lr){6-7}
&  \scalebox{0.9}{APT-T} & \scalebox{0.9}{APD-T} & \scalebox{0.9}{APT-V} & \scalebox{0.9}{APD-V} & \scalebox{0.9}{APT-VL} & \scalebox{0.9}{APD-VL(Ours)} \\
\midrule
Imagenet    & 1.5 & 1.6\textsuperscript{*} & 27.0 & 27.8\textsuperscript{*} & 27.5 & 29.4\textsuperscript{*}\bestglobal \\
Caltech101  & 23.8 & 23.9\textsuperscript{*} & 75.2 & 77.1\textsuperscript{*} & 78.7 & 80.1\textsuperscript{*}\bestglobal \\
DTD         & 6.4 & 7.0\textsuperscript{*} & 36.5 & 41.4\textsuperscript{*} & 38.1 & 44.7\textsuperscript{*}\bestglobal \\
EuroSAT     & 0.2\textsuperscript{*} & 0.2\textsuperscript{*} & 27.6 & 31.0\textsuperscript{*} & 30.8 & 47.3\textsuperscript{*}\bestglobal \\
AirCraft    & 0.7\textsuperscript{*}  & 0.5& 21.9 \textsuperscript{*} & 18.7 & 24.2 & 25.4\textsuperscript{*}\bestglobal \\
Flowers     & 2.6 & 3.3\textsuperscript{*} & 72.5 & 73.0\textsuperscript{*} & 76.8 & 79.0\textsuperscript{*}\bestglobal \\
Food101     & 0.6\textsuperscript{*} & 0.6\textsuperscript{*} & 23.0 & 32.4\textsuperscript{*} & 27.8 & 33.7\textsuperscript{*}\bestglobal \\
OxfordPets  & 2.7 & 3.2\textsuperscript{*} & 50.9 & 54.1\textsuperscript{*} & 52.6 & 56.5\textsuperscript{*}\bestglobal \\
Cars        & 0.6 & 0.7\textsuperscript{*} & 34.3\textsuperscript{*} & 32.9 & 46.6 & 48.2\textsuperscript{*}\bestglobal \\
SUN397      & 1.4 & 1.5\textsuperscript{*} & 31.8 & 32.9\textsuperscript{*} & 34.5 & 37.8\textsuperscript{*}\bestglobal \\
UCF101      & 2.0 & 2.7\textsuperscript{*} & 36.2 & 44.3\textsuperscript{*} & 44.0 & 50.1\textsuperscript{*}\bestglobal \\
\midrule
Avg         & 3.86 & 4.11 \textsuperscript{*} & 39.72 & 42.33\textsuperscript{*} & 43.78 & 48.38\textsuperscript{*}\bestglobal \\
\bottomrule
\end{tabular}
\label{tab: uni_bi}
\vspace{-0.45cm}
}
\end{table}

\begin{figure}
    \centering
    \includegraphics[width=\columnwidth]{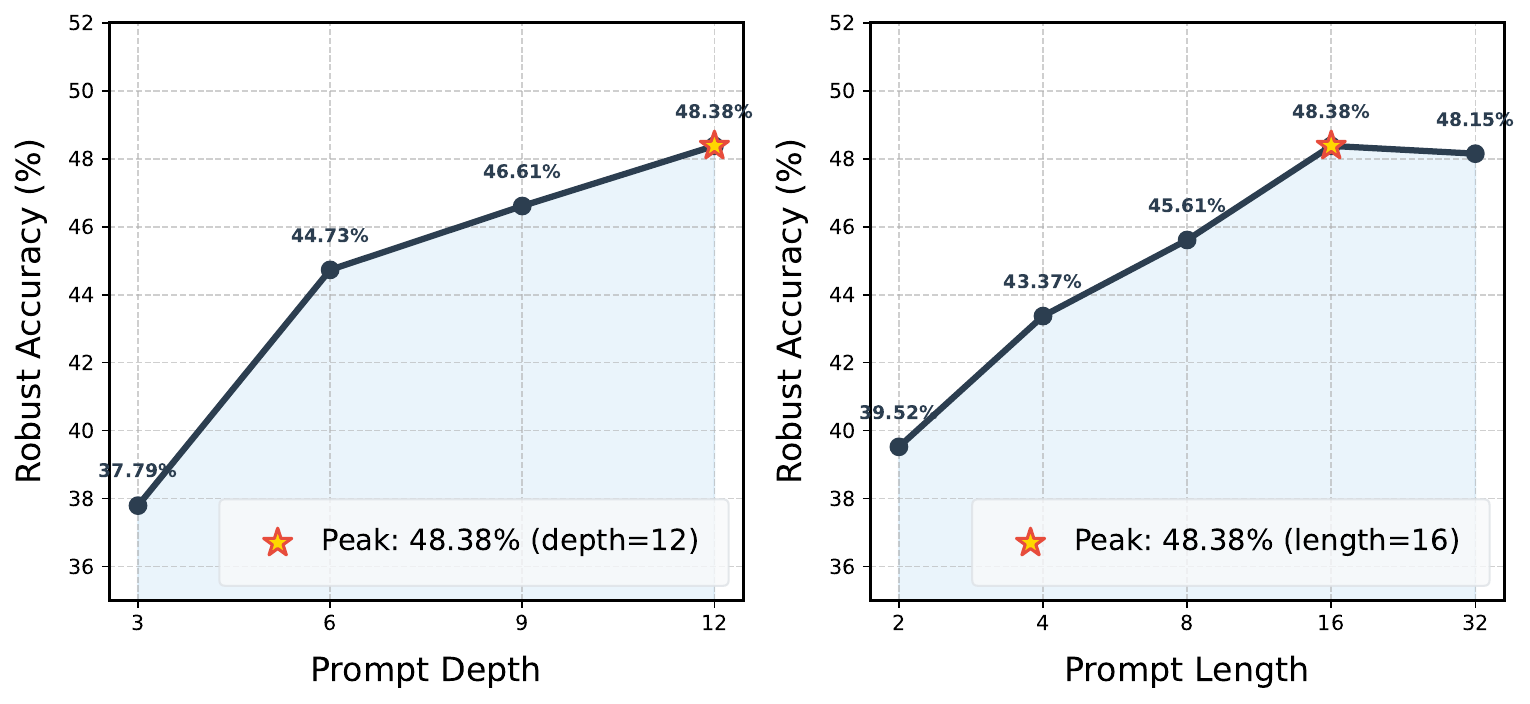}
    \caption{The PGD-100 robustness of our APD  with varying prompt depths \emph{\textbf{(left)}} and lengths \emph{\textbf{(right)}}. The results are averaged over the 11 tested datasets.}
    \label{fig:depth_length}

\end{figure}

\vspace{0.1cm}  
\noindent\textbf{Prompt Depth and Length}
In Figure \ref{fig:depth_length} (left), we examine the impact of prompt depth on the performance of our APD method, evaluated on CLIP with a ViT-B/16 vision backbone. As prompt depth increases, APD’s adversarial robustness steadily improves, suggesting that deeper prompts enhance the model’s ability to resist adversarial attacks by providing richer, more nuanced representations, allowing the model to better capture complex features and patterns. Figure \ref{fig:depth_length} (right) illustrates the effect of prompt length on APD’s performance. Notably, as prompt length increases, APD’s adversarial robustness initially improves, peaking at a length of 16, before declining. At a length of 32, APD exhibits reduced adversarial robustness. This observation suggests that in few-shot scenarios with limited training data, excessively long prompts may lead to underfitting, as additional prompt tokens require more data for effective optimization.                                     

\begin{figure}
    \centering
    \includegraphics[width=\columnwidth]{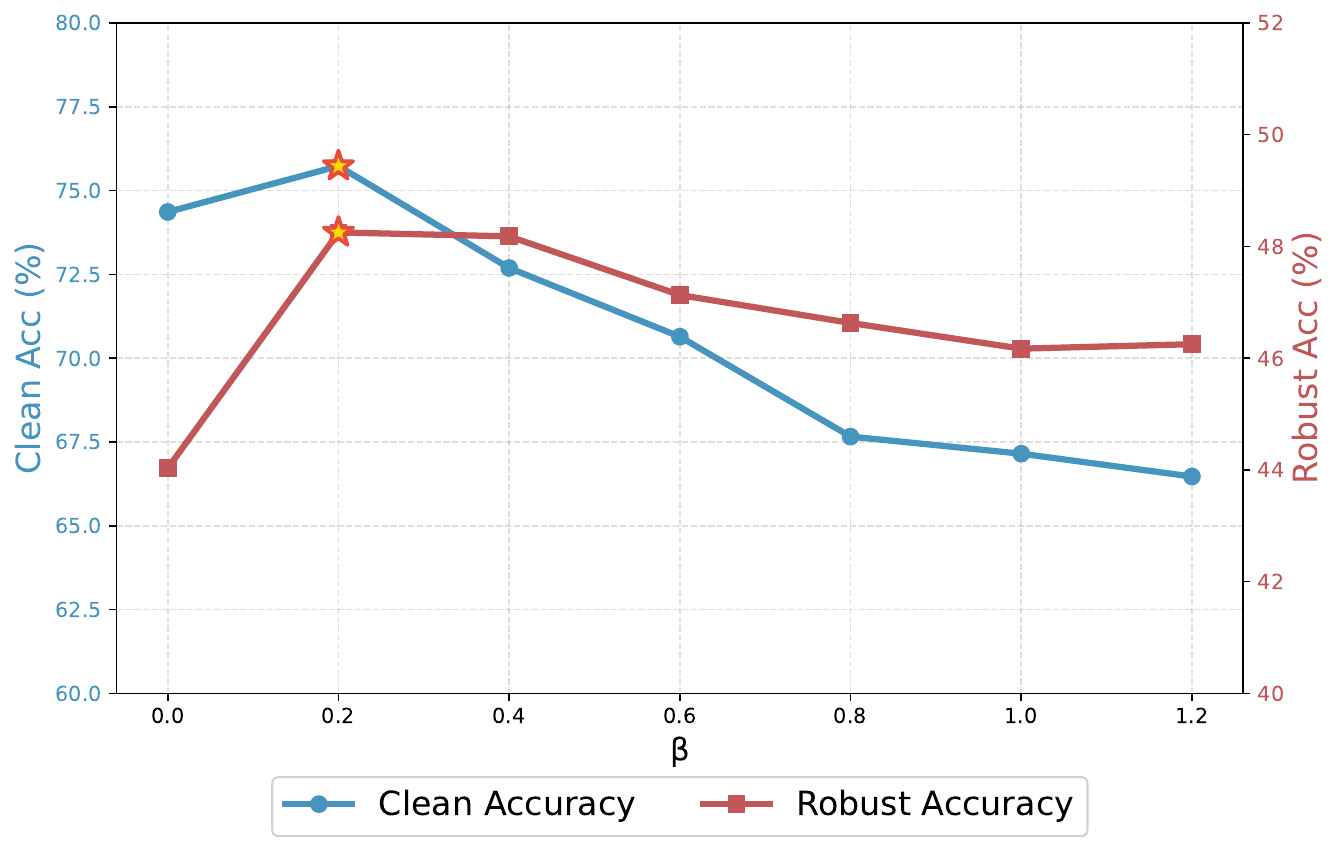}
    \caption{The clean accuracy  and PGD-100 robustness of APD with varying hyperparameter $\beta \in [0, 1.2]$. The results are averaged over the 11 tested datasets. }
    \vspace{-0.45cm}
    \label{fig:ablation_for_beta}
\end{figure}

\vspace{0.1cm}  
\noindent\textbf{Trade-off Between Clean Accuracy and Adversarial Robustness}
We examine the influence of the hyperparameter $\beta$ on the distillation outcomes for APD, using a CLIP model with ViT-B/16 as the vision backbone. As shown in Figure \ref{fig:ablation_for_beta}, increasing $\beta$ results in a gradual decline in the student model's clean accuracy. On the other hand, the robust accuracy initially improves, peaking at $\beta=0.2$ before declining. This pattern arises because, as $\beta$ grows, the teacher model places greater emphasis on feedback from the student model during training, progressively deprioritizing the optimization of the natural objective. As a result, the soft labels produced by the teacher contain less semantic information relevant to generalization on natural examples, ultimately causing a reduction in the student model’s clean accuracy.
Additionally, when $\beta$ exceeds 0.4, the teacher model becomes overly focused on the student’s feedback, leading to weaker performance on natural examples. In this scenario, the teacher’s soft labels become unreliable, causing not only a significant decrease in the student model’s clean accuracy but also a drop in robust accuracy.

\begin{figure}
    \centering
    \includegraphics[width=\columnwidth]{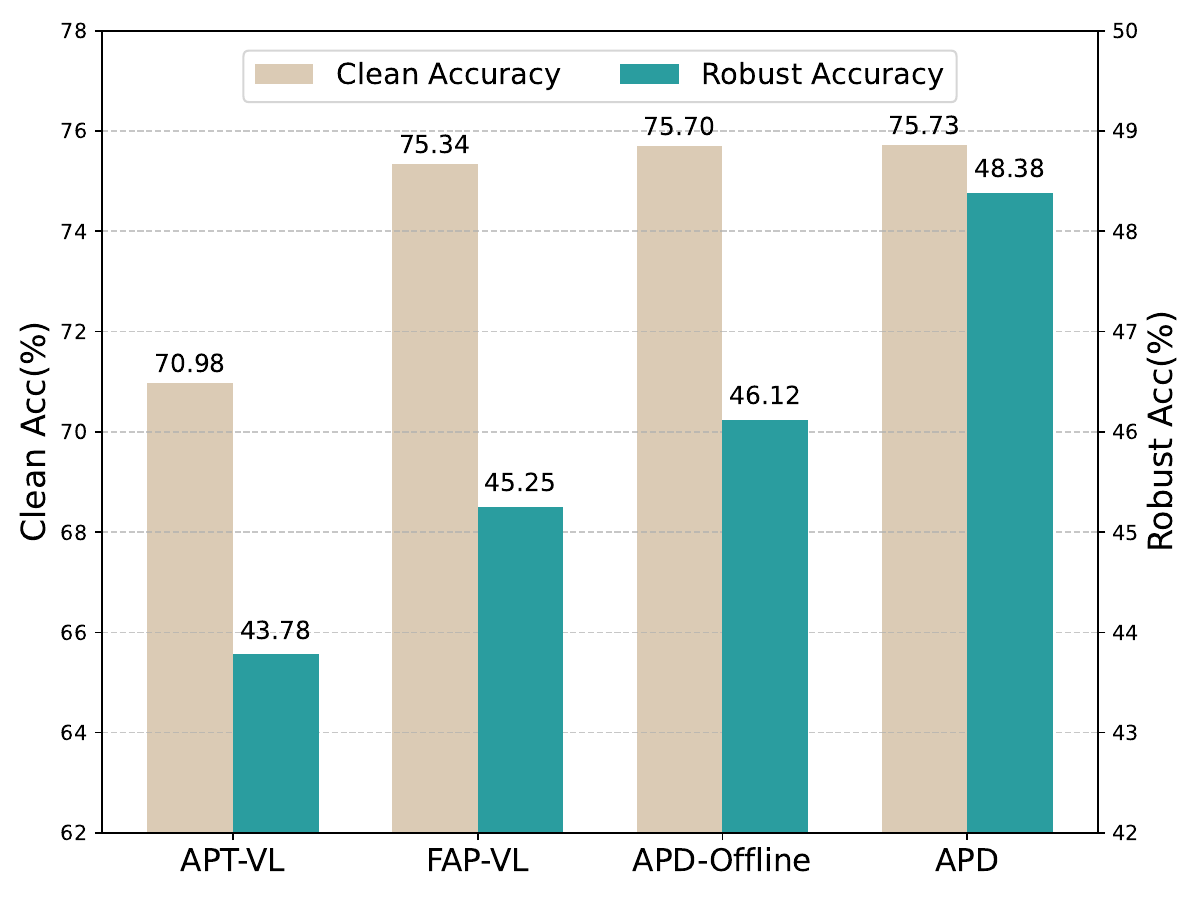}
    \caption{The clean accuracy and PGD-100 robustness of APT-VL, FAP-VL, APD-Offline, and our APD, averaged over the 11 tested datasets, evaluated on CLIP with ViT-B/16 as the vision backbone.}
    \vspace{-0.45cm}
    \label{fig:compared_with_offline_APD}

\end{figure}

\vspace{0.1cm}  
\noindent\textbf{Online APD vs. Offline APD}
Since our APD method is an online distillation method, here we further test an offline version of APD that does not update the teacher model during the distillation process.
For offline APD (denoted as 'APD-Offline'), we first fine-tune the teacher on natural images. During the subsequent distillation process, the student model aligns its outputs with those of the pre-tuned teacher. 
Figure \ref{fig:compared_with_offline_APD} presents the comparison results of two bimodal defense baselines (APT-VL and FAP-VL), APD-Offline, and our APD. Notably, APD-Offline outperforms both baseline methods in terms of clean accuracy and PGD-100 robustness, demonstrating the effectiveness of the distillation strategy.
This benefit of using a clean teacher model is maximized by online distillation (i.e., our APD method). 
Our online approach enables the teacher model to be student-aware during distillation, bridging the teacher-student gap and embodying the principle that teachers should instruct students according to their aptitude~\cite{li2022shadow}.

\section{Conclusion}
In this work, we studied the problem of improving the adversarial robustness of non-robust pre-trained Vision-Language Models (VLMs) like CLIP on downstream tasks. To address this challenge, we introduced a novel method \emph{Adversarial Prompt Distillation (APD)}, which leverages both textual and visual prompts to strengthen defense against image-modality adversarial attacks. Additionally, APD combines adversarial prompt tuning (APT) with knowledge distillation, using a cleanly pre-trained teacher CLIP model to distill soft-label guidance into the student model.
Our extensive experiments on multiple benchmark datasets demonstrate that APD significantly improves both natural accuracy and adversarial robustness over current state-of-the-art defense methods across different architectures. 
Our work validates the potential of using a non-robust teacher model to improve VLM generalization and adversarial resistance.

{
    \small
    \bibliographystyle{ieeenat_fullname}
    \bibliography{main}
}

\end{document}